%% file: main.tex
\documentclass[]{xiaomi}
\usepackage{amsmath,amsfonts,amssymb}

\usepackage{colortbl}
\usepackage{makecell}
\usepackage{tabularx}
\usepackage{longtable}

\usepackage{adjustbox}
\usepackage{enumitem}
\usepackage{xspace}
\usepackage{pifont}
\usepackage{fvextra}
\usepackage{nicefrac}

\ifPDFTeX\else
\fi

\definecolor{BrickRed}{rgb}{.72,0,0}
\definecolor{darkgreen}{rgb}{0.0,0.5,0.0}
\definecolor{ForestGreen}{RGB}{34,139,34}
\definecolor{LakeBlue}{RGB}{0,61,153}
\definecolor{MiOrange}{RGB}{255,225,204}
\definecolor{xiaomiorange}{RGB}{255,103,0}
\definecolor{Orange}{RGB}{255,224,199}
\definecolor{Hex}{RGB}{225,213,231}
\definecolor{Gray}{gray}{0.85}

\newcommand{\METHODNAME}{Xiaomi-OCR-0\xspace}
\newcommand{\METHODNAMEhl}{{\sffamily\bfseries\textcolor{xiaomiorange}{Xiaomi-OCR-0}}\xspace}
\newcommand{\hfmark}{%
  \raisebox{-0.15ex}{\includegraphics[height=0.95em]{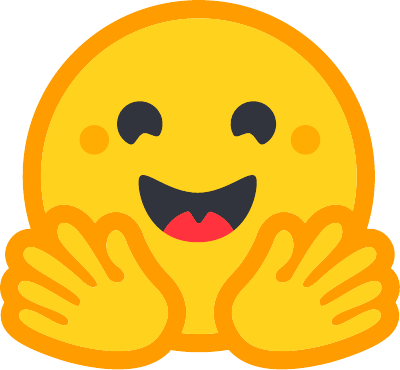}}%
  \hspace{0.22em}}
\providecommand{\OverviewFigure}{figures/overview_classic.pdf}

\newtcolorbox{takeaway}{
  enhanced,
  colback=xiaomiorange!4!white,
  colframe=xiaomiorange,
  boxrule=0.5pt,
  borderline west={2pt}{0pt}{xiaomiorange},
  arc=1.5mm,
  left=3mm, right=3mm, top=2mm, bottom=2mm,
  before skip=10pt, after skip=10pt
}

\titleformat{\paragraph}[runin]
  {\normalfont\normalsize\bfseries\color[HTML]{FF7E00}}
  {}{0pt}{}
\titlespacing*{\paragraph}{0pt}{6pt}{1em}

\title{\METHODNAMEhl Technical Report}
\author{SeerRay Team}
\contribution{See \hyperref[sec:contributions]{Contributions and Acknowledgments} section for a full author list.}

\abstract{\input{sections/abstract}}

\begin{document}
\maketitle
\vspace{-5.3mm}
\noindent\begin{minipage}{\linewidth}
\centering
\vspace{-10pt}

\includegraphics[width=0.8\linewidth]{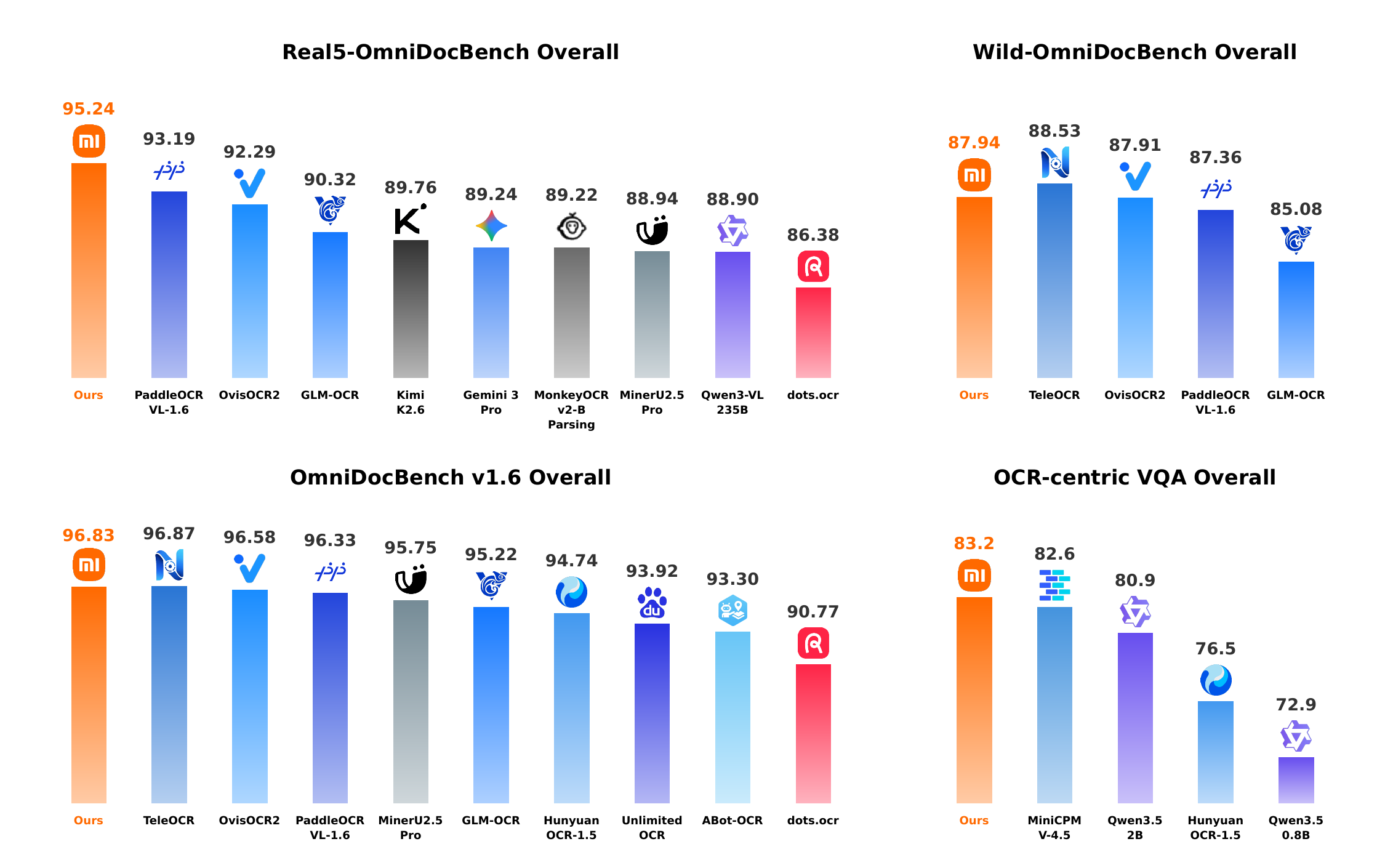}
\captionof{figure}{Performance overview of \METHODNAME on acquisition robustness (top left), in-the-wild parsing (top right), standard document parsing (bottom left), and OCR-centric understanding (bottom right). 
}
\label{fig:performance_overview}
\end{minipage}
\clearpage
\tableofcontents
\newpage

\input{sections/introduction}
\input{sections/data_engine}
\input{sections/training_recipe}
\input{sections/evaluation}
\input{sections/analysis}
\input{sections/conclusion}

\bibliographystyle{plainnat}
\bibliography{ref}

\input{sections/contributions}

\end{document}

%% file: sections/abstract.tex
Compact OCR-specific vision-language models achieve strong document parsing performance, but often rely on costly supervision and focus primarily on visual-text reconstruction. We introduce \METHODNAMEhl, a unified 0.8B model for document parsing and OCR-centric understanding. We build an approximately 170M-sample OCR-centric corpus using an automated data engine that combines expert consensus, render-based verification, and targeted synthesis. Starting from Qwen3.5-0.8B, our progressive training recipe combines Q-Mask-based text anchoring, continued pretraining, and mixed-task reinforcement learning (Mix-RL). \METHODNAME achieves 95.24 on Real5-OmniDocBench, 96.83 on OmniDocBench v1.6, and 87.94 on Wild-OmniDocBench, while reaching an average score of 83.2 across five OCR-oriented VQA benchmarks. Ablations further show that, with sufficient parsing training, OCR-centric understanding supervision provides additional gains for document parsing.

\hfmark Homepage: \url{https://huggingface.co/spaces/SeerRay-Lab/Xiaomi-OCR-0}.

%% file: sections/introduction.tex
\section{Introduction}

Optical Character Recognition (OCR)~\citep{subramani2020survey,faizullah2023survey,fu2025multimodal} connects text-rich visual inputs to language-based systems. Modern OCR systems increasingly need to support two complementary capabilities. Document parsing reconstructs a document's content and structure, while OCR-centric understanding selects and interprets task-relevant visual text~\citep{xu2020layoutlm,kim2022ocrfree,singh2019towards,mathew2021docvqa,masry2022chartqa}. We use OCR-centric understanding to encompass OCR-oriented visual question answering (OCR-VQA) and key information extraction (KIE). These capabilities serve different purposes but are closely related in practice: parsing provides a faithful and structured representation of the document. At the same time, understanding enables selective, task-conditioned reasoning over its visual text. Their shared need for accurate recognition and layout interpretation motivates a single model that supports both through task-specific instructions. Recent OCR-specific vision-language models (VLMs) demonstrate strong parsing with compact and efficient architectures~\citep{wei2026deepseekocr,zhang2026paddleocrvl16,wang2026mineru25pro,cai2026navidc,lu2026ovisocr2}. Extending this efficiency to OCR-centric understanding requires models to retain fine-grained text recognition while learning to select and interpret visual text according to the task.

Jointly developing these capabilities presents two challenges. First, reliable supervision is expensive to scale. Text, tables, formulas, and reading order require different verification procedures, while plausible model-generated annotations can conceal subtle errors. Second, their objectives differ: parsing requires faithful, exhaustive reconstruction, whereas understanding requires selective, task-conditioned responses. Simply mixing their training data therefore does not guarantee positive transfer. Our experiments reveal that the interaction between parsing and understanding depends strongly on the maturity of the underlying parsing capability. When parsing capability is still immature, introducing understanding supervision too early can interfere with developing a strong parsing foundation, resulting in negative transfer. As parsing capability matures, however, the same understanding supervision becomes beneficial, shifting from negative transfer.

We introduce \METHODNAME, a unified 0.8B VLM initialized from Qwen3.5-0.8B~\citep{qwen2026qwen35}. To support large-scale OCR-centric training, we first build a high-quality data engine that integrates automated annotation, verification, sample mining, and targeted synthesis, enabling diverse, reliable supervision at scale with limited human intervention. Using the resulting ~170M-sample corpus, we train \METHODNAME with a progressive recipe that moves from Q-Mask-based text anchoring~\citep{xu2026qmask} to multi-task continued pretraining (CPT), followed by mixed-task reinforcement learning (Mix-RL) on hard examples with verifiable rewards. The resulting model jointly supports document parsing and OCR-centric understanding through a shared instruction interface. 

We also compare Mix-RL with multi-teacher on-policy distillation (MOPD)~\citep{ma2026mopd}, showing that MOPD reaches competitive parsing performance with less training compute, while a longer Mix-RL run achieves higher final performance on both parsing and understanding. We further use task-specific 4B teachers to examine the relative capacity demands of the two capabilities.

With PP-DocLayoutV3~\citep{cui2026rtdoclayoutrealtimeendtoenddocument} providing layout detection, \METHODNAME achieves an Overall score of 95.24 on Real5-OmniDocBench, the highest among the compared methods. It scores 96.83 on OmniDocBench v1.6 and 87.94 on Wild-OmniDocBench, showing strong parsing on standard and recaptured documents. For OCR-centric understanding, it obtains a mean score of 83.2 across DocVQA, InfoVQA, ChartQA, OCRBench, and TextVQA, and achieves 59.09\% accuracy on our in-house KIE benchmark.

\begin{samepage}
Our contributions are threefold:
\begin{itemize}
    \item We identify a capability-dependent interaction between document parsing and OCR-centric understanding, showing that understanding supervision shifts from negative to positive transfer as the underlying parsing capability matures.
    \item We build a scalable automated data pipeline that enables large-scale high-quality supervision with minimal human intervention.
    \item We develop \METHODNAME, a unified 0.8B VLM that performs strongly across document parsing, OCR-VQA, and KIE.
\end{itemize}
\end{samepage}
\clearpage

%% file: sections/data_engine.tex
\section{Data Engine}
\label{sec:data_engine}

We construct an OCR-centric training corpus of approximately 170M samples spanning document parsing and OCR-centric understanding. The corpus uses a unified image--instruction--response format. This section focuses on the document-parsing data engine, which provides supervision for text, tables, and formulas. OCR-centric understanding supervision combines existing OCR-VQA data with structured field annotations generated according to task-specific schemas for KIE.

\paragraph{Region-level and page-level parsing data.}
\label{sec:parsing_data_granularity}
A region-level sample pairs a cropped text, table, or formula region with its recognition target. A page-level sample pairs a complete document image with its full parsing target, including content, structure, and reading order. Region-level supervision trains recognition on crops used in a two-stage pipeline; page-level supervision trains direct end-to-end parsing from a full-page image. We train the same VLM with both forms of supervision. These terms describe training inputs, and the reported data counts refer to samples rather than uniformly to document pages.

The data engine comprises three interacting components that address label selection, training value, and distributional coverage. The \emph{automated annotation pipeline} (Section~\ref{sec:annotation_pipeline}) selects candidate labels for training through heterogeneous-expert consensus and render-guided refinement. \emph{Multi-factor sample mining} (Section~\ref{sec:sample_mining}) determines which samples to emphasize, combining expert agreement, model self-consistency, and semantic clustering into training mixtures. The \emph{synthetic data generation pipeline} (Section~\ref{sec:synthetic_data_generation_pipeline}) extends the corpus where real samples are insufficient, using coverage gaps and verified failure patterns as generation targets.

\subsection{Automated Annotation Pipeline}
\label{sec:annotation_pipeline}

\begin{figure*}[t]
    \centering
    \includegraphics[width=\textwidth]{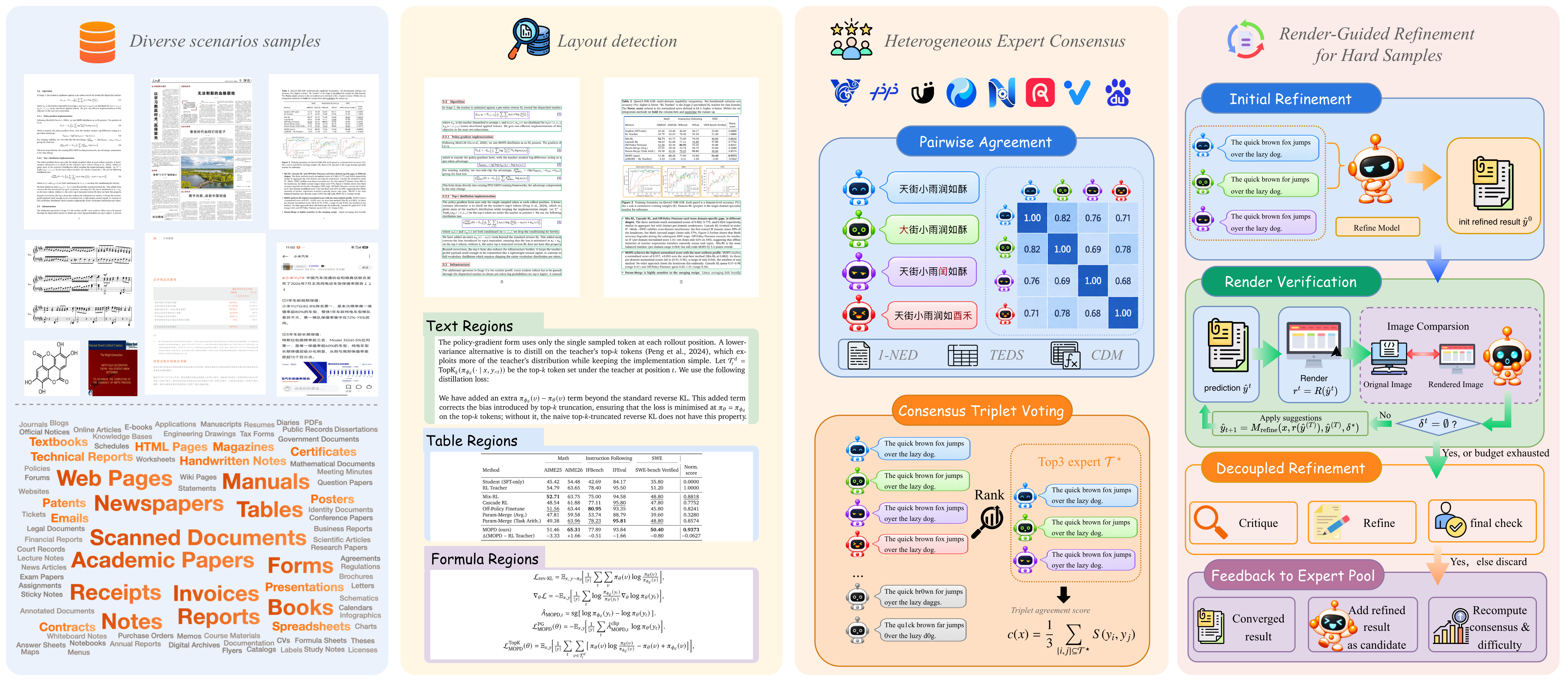}
    \caption{Automated annotation pipeline. Heterogeneous experts annotate document regions, triplet consensus scores them, and render-based verification refines them when agreement is insufficient. Refined candidates return to the expert pool for re-consensus.}
    \label{fig:data_engine}
\end{figure*}

The annotation pipeline operates on document regions, as illustrated in Figure~\ref{fig:data_engine}. PP-DocLayoutV3~\citep{cui2026rtdoclayoutrealtimeendtoenddocument} first detects semantic regions on each page. We retain text, table, and formula regions, crop them individually, and annotate each crop with a heterogeneous expert pool. Regions with high expert agreement are accepted directly; regions with lower agreement undergo render-guided correction before being re-evaluated by the pool. Selected unresolved cases receive targeted human verification rather than entering training with uncertain labels.

\subsubsection{Ensemble Triplet Consensus}
\label{sec:ETC}

A single parsing model can produce plausible but incorrect transcriptions or structures, making its predictions an insufficient quality signal. Agreement across heterogeneous models provides a practical alternative, although shared errors can still occur. We therefore introduce \emph{Ensemble Triplet Consensus} (ETC), which ranks candidates using pool-wide agreement, selects a three-expert subset, and uses agreement within that subset to score annotations and route uncertain samples.

ETC differs from two related consensus schemes. CMCV in MinerU2.5-Pro~\citep{wang2026mineru25pro} grades samples using the agreement pattern of three models relative to the target model. MCV in TeleOCR~\citep{cai2026navidc} selects the prediction with the highest pool-wide agreement and applies a binary acceptance threshold. ETC instead separates candidate ranking from annotation scoring and supports graded routing without requiring the target model to belong to the expert pool.

\paragraph{Expert pool and pairwise agreement.}
For a region $x$, each expert in $\mathcal{M}=\{M_1,\ldots,M_N\}$ independently produces a prediction $y_i=M_i(x)$. The initial pool contains eight models with different architectures, training data, and parsing strategies: PaddleOCR-VL-1.6~\citep{zhang2026paddleocrvl16}, MinerU2.5-Pro~\citep{wang2026mineru25pro}, GLM-OCR~\citep{duan2026glmocr}, dots.ocr~\citep{li2025dotsocr}, HunyuanOCR-1.5~\citep{li2026hunyuanocr15}, TeleOCR~\citep{cai2026navidc}, OvisOCR2~\citep{lu2026ovisocr2}, and Qianfan-OCR~\citep{dong2026qianfanocr}. Their diversity provides complementary predictions rather than repeated estimates from a single model.

We measure pairwise agreement with a region-specific similarity $S(y_i,y_j)\in[0,1]$: the complement of normalized edit distance (NED), $1-\mathrm{NED}$, for text; Tree-Edit-Distance-based Similarity (TEDS)~\citep{pubtabnet} for tables; and Character Detection Matching (CDM)~\citep{cdm} for formulas. The same similarity is used for candidate ranking and triplet scoring within each region type.

\paragraph{Triplet selection and pseudo-label construction.}
We first compute each expert's mean agreement with the rest of the pool:
\begin{equation}
    C_i=\frac{1}{N-1}\sum_{j\neq i}S(y_i,y_j).
\end{equation}
Let $C_{(1)}\geq C_{(2)}\geq\cdots\geq C_{(N)}$ denote the ranked scores. The indices of the three highest-scoring experts form the consensus triplet $\mathcal{T}^{\star}=\{(1),(2),(3)\}$. This is a pool-wide ranking rule, rather than an exhaustive search for the triplet with maximum internal agreement. We define the region's \emph{expert agreement score} as
\begin{equation}
    a(x)=\frac{1}{3}\sum_{\substack{i,j\in\mathcal{T}^{\star}\\i<j}}S(y_i,y_j).
    \label{eq:expert_agreement}
\end{equation}
This score measures agreement within the selected triplet; it is not a calibrated probability that the pseudo-label is correct. We select as the pseudo-label the triplet medoid, i.e., the candidate with the highest total agreement with the other two members:
\begin{equation}
    y^{\star}(x)=y_{i^{\star}},
    \qquad
    i^{\star}=\arg\max_{i\in\mathcal{T}^{\star}}
    \sum_{j\in\mathcal{T}^{\star}\setminus\{i\}}S(y_i,y_j).
\end{equation}
Pool-wide scores thus determine which candidates participate, while $a(x)$ records their mean pairwise agreement.

\paragraph{Agreement tiers and routing.}
We divide regions into three agreement tiers:
\begin{equation}
    d_{\mathrm{agree}}(x)=
    \begin{cases}
        \text{high}, & a(x)\geq 0.9,\\[2pt]
        \text{medium}, & 0.6\leq a(x)<0.9,\\[2pt]
        \text{low}, & a(x)<0.6.
    \end{cases}
    \label{eq:agreement_tier}
\end{equation}
Regions in the high-agreement tier are accepted directly. Medium- and low-agreement regions are sent to the refine-and-judge module in Section~\ref{sec:refine_judge}. Agreement tiers measure consensus across expert annotators. During sample mining, we also measure model self-consistency across repeated predictions for each sample.

\subsubsection{Render-Guided Refine-and-Judge}
\label{sec:refine_judge}

When experts disagree, comparing their output strings alone may not reveal whether a table topology or formula structure matches the source image. We therefore render a candidate annotation back into an image and compare it with the original region. This makes structural discrepancies visually explicit and follows the render-then-verify paradigm used in recent OCR systems~\citep{wang2026mineru25pro,zhang2026paddleocrvl16,cai2026navidc}.

We use Qwen3.5-122B-A10B~\citep{qwen2026qwen35} as the refinement model $M_{\mathrm{refine}}$. It is outside the initial expert pool and handles visual verification and iterative correction, while the task-specialized experts provide recognition candidates.

\paragraph{Consensus-conditioned initialization.}
Within $\mathcal{T}^{\star}$, we select the pair $(a,b)$ with the highest mutual agreement and provide their predictions, together with their source identifiers, as references:
\begin{equation}
    \hat{y}^{(0)}=M_{\mathrm{refine}}\big(x,\{(a,y_a),(b,y_b)\}\big).
\end{equation}
The model analyzes the discrepancies against the source image and produces an initial correction. Restricting the references to the most consistent pair limits conflicting candidate information without treating either reference as ground truth.

\paragraph{Iterative render verification.}
The initial refinement stage has a verification budget of $T=3$ rounds. At round $t\in\{0,\ldots, T-1\}$, we render the current prediction as $r^{(t)}=R(\hat{y}^{(t)})$, using HTML for text and tables and LaTeX for formulas. The refinement model compares the source and rendered images and returns edit suggestions:
\begin{equation}
    \delta^{(t)}=M_{\mathrm{refine}}\big(x,r^{(t)},\hat{y}^{(t)}\big).
\end{equation}
If $\delta^{(t)}=\varnothing$, verification terminates. Otherwise, a separate call applies the suggestions and returns a complete corrected annotation:
\begin{equation}
    \hat{y}^{(t+1)}=M_{\mathrm{refine}}\big(x,r^{(t)},\hat{y}^{(t)},\delta^{(t)}\big).
\end{equation}
After a correction, a task-specific correction-magnitude threshold also permits early stopping when successive predictions change only slightly. Either stopping condition sends the resulting candidate directly to re-consensus; only samples that remain unconverged after exhausting the verification budget enter the second stage below. Expert references are supplied only during initialization; later rounds remain grounded in the source image, the current prediction, and its rendering.

\paragraph{History-aware decoupled refinement.}
If the first stage exhausts its verification budget without convergence, we apply a second-stage Critique--Refine--Verify procedure. The key difference is a history-aware diagnosis: the critique call receives the full first-stage verification history $\mathcal{H}$ to identify errors that persist across rounds. Diagnosis, correction, and final verification are performed in three separate calls:
\begin{equation}
    \begin{aligned}
        \delta^{\star} &= M_{\mathrm{refine}}\big(x,R(\hat{y}^{(T)}),\hat{y}^{(T)},\mathcal{H}\big),\\
        \hat{y}_2 &= M_{\mathrm{refine}}\big(x,R(\hat{y}^{(T)}),\hat{y}^{(T)},\delta^{\star}\big),\\
        \delta_2 &= M_{\mathrm{refine}}\big(x,R(\hat{y}_2),\hat{y}_2\big).
    \end{aligned}
\end{equation}
The second-stage candidate proceeds only when the final check returns no further edits, i.e., $\delta_2=\varnothing$.

\paragraph{Re-consensus and residual review.}
A converged refined prediction is added to the original predictions as one additional candidate. ETC then recomputes the ranking, triplet, pseudo-label, and agreement score. Only regions assigned to the high-agreement tier by Eq.~\eqref{eq:agreement_tier} after re-consensus enter the automatically accepted corpus; refinement is not assumed to increase agreement. Unresolved regions are withheld, with selected cases routed to human verification. Verified hard cases can also provide reference patterns for synthesis.

\subsection{Multi-Factor Sample Mining}
\label{sec:sample_mining}

High expert agreement is useful for label selection but does not make every sample equally useful for training. Sampling only by corpus frequency can underrepresent rare layouts, whereas sampling only by low self-consistency can overconcentrate on a narrow set of patterns. We therefore combine expert agreement, model self-consistency, and semantic coverage to construct task-specific mixtures for continued pretraining (CPT) and reinforcement learning (RL). The same signals identify where synthetic data are needed.

For each region sample $x_i$, we compute:
\begin{itemize}
    \item \textbf{Expert agreement} $a_i\in[0,1]$, given by the score in Eq.~\eqref{eq:expert_agreement}. It measures agreement among the selected experts.
    \item \textbf{Self-consistency} $u_i\in[0,1]$, given by the mean pairwise similarity across multiple stochastic inference passes of the current model, using the same region-specific similarity $S$ as in Section~\ref{sec:ETC}. Lower self-consistency indicates greater variation across the model's predictions for the same input.
    \item \textbf{Semantic cluster} $k(i)\in\{1,\ldots,K\}$, obtained by extracting region-image embeddings with Qwen3-VL-Embedding~\citep{qwen3vlembedding} and applying K-Means separately for each task type. Clusters capture variation in visual layout and semantic content.
\end{itemize}

The model used to compute $u_i$ is not part of the expert pool used to compute $a_i$. These scores address different questions: how strongly the experts agree on a label and how consistently the model responds to the sample. Repeated predictions can agree while being incorrect, so self-consistency measures prediction stability. We use it with expert agreement to select candidates for error analysis.

\paragraph{Cluster-based sampling.}
Sampling proceeds in two stages. First, within each task type, let $\mathcal{C}_1,\ldots,\mathcal{C}_K$ denote the nonempty clusters and let $n\geq K$ be the nominal sampling budget. We allocate cluster-level quotas to retain broad distributional coverage: large clusters receive quotas proportional to their size, while small clusters ($|\mathcal{C}_k|<n/K$) are amplified by an over-sampling factor $\rho$:
\begin{equation}
    q_k=
    \begin{cases}
        \max\left(1,\lfloor \rho\cdot|\mathcal{C}_k|\rfloor\right), & |\mathcal{C}_k| < n/K, \\[6pt]
        \min\left(|\mathcal{C}_k|,\max\left(1,\operatorname{round}\left(n\cdot|\mathcal{C}_k|/N_{\mathrm{data}}\right)\right)\right), & |\mathcal{C}_k| \geq n/K,
    \end{cases}
\end{equation}
where $N_{\mathrm{data}}$ is the total number of samples. This heuristic does not, in general, make the quotas sum to $n$; the number of draws is $\sum_k q_k$.

Second, we draw $q_k$ samples with replacement within each cluster using weights based on expert agreement and model self-consistency. Repeated draws allow rare clusters to receive quotas larger than their number of distinct samples. For $x_i\in\mathcal{C}_k$,
\begin{equation}
    P(i\mid k)=
    \frac{(a_i+\varepsilon)^{f_a/\tau}(1-u_i+\varepsilon)^{f_u/\tau}}
    {\displaystyle\sum_{j:x_j\in\mathcal{C}_k}(a_j+\varepsilon)^{f_a/\tau}(1-u_j+\varepsilon)^{f_u/\tau}},
    \label{eq:within_cluster_sampling}
\end{equation}
where $f_a,f_u\geq 0$ control the emphasis on expert agreement and model self-consistency, $\varepsilon>0$ keeps both bases strictly positive, and $\tau>0$ is the sampling temperature. At $\tau=1$, the weights are proportional to $(a_i+\varepsilon)^{f_a}(1-u_i+\varepsilon)^{f_u}$; as $\tau\to\infty$, sampling approaches uniform selection within each cluster. Adjusting these parameters supports broad-coverage mixtures for CPT and mixtures that place greater weight on low-self-consistency samples for RL. At the same time, cluster quotas preserve representation across the data distribution.

\paragraph{Hard-sample mining.}
We select candidate samples with high expert agreement and low model self-consistency. Using $a_i$ and $u_i$ defined above, the candidate pool is
\begin{equation}
    \mathcal{D}_{\mathrm{candidate}}=
    \left\{x_i\in\mathcal{D}\;\middle|\;
    a_i\geq\tau_a,\ u_i\leq\tau_u\right\},
    \label{eq:hard_case_pool}
\end{equation}
where $\tau_a$ and $\tau_u$ are thresholds for expert agreement and model self-consistency, respectively. We verify candidate errors by comparing model predictions with verified reference annotations, then categorize the confirmed discrepancies.

The agent combines cluster assignments with visual and structural attributes, task types, and discrepancies from reference annotations to categorize recurring failure patterns.

\subsection{Synthetic Data Generation}
\label{sec:synthetic_data_generation_pipeline}

Sample mining can redistribute existing examples, but it cannot supply missing structures, writing styles, or character combinations. The synthetic pipeline complements real-data annotation by generating image--annotation pairs in two modes: \emph{coverage-driven synthesis} expands underrepresented parts of the data distribution, while \emph{failure-driven synthesis} creates targeted variants around diagnosed model errors.

Agents automate material organization, generator construction, and recipe execution; human experts define the initial rules, constraints, and reference materials, and review the resulting data and evaluation outcomes. When recipe validation involves training, agents also orchestrate data mixing, supervised fine-tuning (SFT), evaluation, and recipe revision.

\paragraph{Shared synthesis infrastructure.}
Both modes use reusable generation and quality-control components; they differ in how they select generation targets and synthesis parameters. Drawing on published synthesis methods and an internally curated collection of fonts, text corpora, and rendering assets authorized for this use, the agent composes recipes from structural templates, text, visual styles, backgrounds, and degradation parameters. For template-based documents, structurally annotated HTML/CSS pages are rendered into image--annotation pairs using a browser or another rendering engine. Quality control combines task-appropriate rule-based validation, rendering-consistency checks, deduplication, and filtering to verify agreement among the image, annotation, output structure, and task format. Invalid samples are rejected or regenerated before training.

\subsubsection{Coverage-Driven Synthesis}
\label{sec:coverage_driven_synthesis}

Coverage-driven synthesis addresses gaps in content, structure, visual style, and domain, without requiring an observed model error on every target pattern. We use template-driven generation for conventional document structures and material-driven simulation for low-resource scenarios.

\paragraph{Conventional document structures.}
For text, tables, formulas, and common document layouts, the agent builds a template seed library from two sources: underrepresented clusters identified by sample mining and existing diverse templates. The latter include long tables, complex headers, merged cells, and uncommon page layouts. Cluster analysis supplies evidence of coverage gaps, while the template collection provides structures that may be rare or absent in the real-data pool. These seeds configure the shared generators to vary content, appearance, and degradation conditions around the selected coverage targets.

\paragraph{Low-resource visual-text scenarios.}
For calligraphy, special fonts, and low-resource scenarios, real examples alone may not provide sufficient character and style coverage. We use reference works, font files, individual characters or partial glyphs, and real backgrounds as visual priors, within the scope of their respective permissions. Guided by the target text corpus and commonly confused characters, the agent organizes, renders, and augments these materials to vary character combinations, writing styles, and degradation conditions. Real samples anchor visual authenticity, while synthetic expansion increases coverage of long-tail characters and combinations.

\subsubsection{Failure-Driven Synthesis}
\label{sec:failure_driven_synthesis}

Failure-driven synthesis takes the candidate pool $\mathcal{D}_{\mathrm{candidate}}$ defined in Eq.~\eqref{eq:hard_case_pool}, diagnoses the current model's errors while training, and generates controlled variants that preserve the difficult factors. Unlike coverage-driven synthesis, its objective addresses an observed model weakness rather than a distributional gap alone.

\paragraph{Error diagnosis.}
Before error analysis or synthesis, we randomly partition the candidate pool by source document into a training set $\mathcal{D}_{\mathrm{candidate}}^{\mathrm{train}}$ and a frozen validation set $\mathcal{D}_{\mathrm{candidate}}^{\mathrm{val}}$. When document identity is unavailable, the source page is the grouping unit. Regions from the same source group, together with linked near-duplicates and synthetic derivatives, remain in the same partition. Validation source groups are excluded not only from the candidate training set but also from the base SFT mixture and all synthesis inputs in every iteration. Only training groups supply examples for error analysis, synthesis conditioning, and SFT. The validation split supports recipe selection, gain measurement, and stopping decisions, rather than serving as an independent final test set.

The agent analyzes the training split by task type, semantic cluster, visual and structural attributes, and discrepancies between model predictions and reference annotations. Comparing against references helps distinguish incorrect predictions from inference instability alone and organize errors into recurring patterns for targeted generation.

\paragraph{Hard-sample-conditioned generation.}
For each diagnosed training example, the pipeline conditions the shared generators on its source image and annotation to reconstruct its content and structure. The agent compares the reconstruction with the original, then adjusts templates, text, fonts, layout, and degradation parameters. It generates multiple variants that retain the relevant difficulty---such as merged cells, nested formulas, complex layouts, or output-format constraints---while varying other aspects of content and appearance. The goal is not merely to duplicate a hard image, but to expand the range of examples exhibiting its diagnosed failure pattern. The variants then pass through the shared quality-control components.

\paragraph{Recipe validation and iteration.}
To evaluate a synthesis recipe, the filtered targeted data are mixed at a predefined ratio with the base training data after excluding all validation source groups and their linked derivatives. This validation-excluded mixture is used for SFT, and the exclusion is maintained across recipe revisions. The resulting model is evaluated on the frozen hard validation set, with regression evaluation used to check for degradation beyond the targeted patterns. Based on these results, the agent accepts the recipe, adjusts synthesis parameters, or revisits the error analysis. Human experts review sampled outputs and final results rather than manually initiating each iteration.

Coverage-driven synthesis expands underrepresented patterns; failure-driven synthesis tests whether targeted additions resolve verified errors.

%% file: sections/training_recipe.tex
\section{Progressive Training Recipe}

\begin{figure}[t]
\centering
\includegraphics[width=\textwidth]{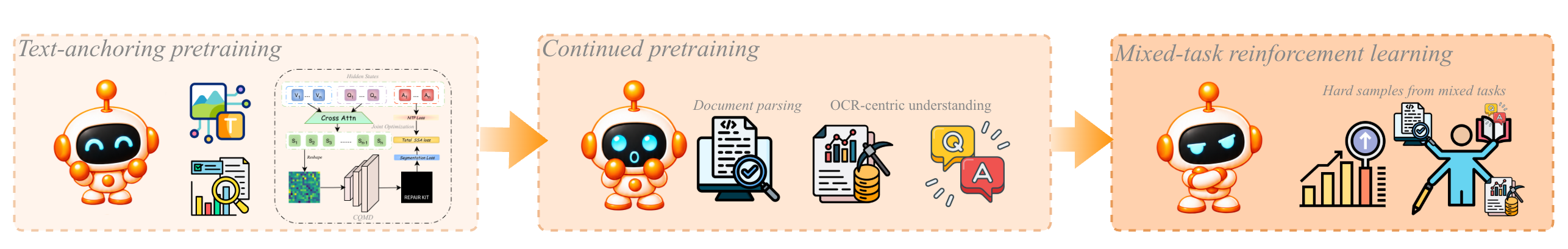}
\caption{Overview of the progressive training recipe.}
\label{fig:training_pipeline}
\end{figure}

Figure~\ref{fig:training_pipeline} illustrates the overall training recipe. Starting from Qwen3.5-0.8B~\cite{qwen2026qwen35}, the model first undergoes text-anchoring pretraining following Q-Mask, which establishes fine-grained alignment between text content and its spatial location. It then undergoes OCR-centric continued pretraining to expand its multi-task OCR capabilities, followed by Mix-RL on hard examples to optimize document parsing and OCR-centric understanding with verifiable task rewards.

\subsection{Text-Anchoring Pretraining}
\label{sec:grounding_cpt}

Text-anchoring pretraining follows the Q-Mask framework~\cite{xu2026qmask}, which introduces a query-conditioned mask decoder and supervises a spatial mask together with the answer tokens. The model is trained with Q-Mask's Spatial Supervision Alignment (SSA) objective, which jointly optimizes next-token prediction and mask prediction:
\begin{equation}
\mathcal{L}_{\mathrm{SSA}} = \lambda_{\mathrm{txt}}\, \mathcal{L}_{\mathrm{NTP}}
+ \lambda_{\mathrm{seg}}\, (\mathcal{L}_{\mathrm{Dice}} + \mathcal{L}_{\mathrm{CE}}),
\end{equation}

The training data combine the original TextAnchor-26M dataset with document parsing data produced by our data engine. Applying the same Q-Mask construction scheme converts these samples into text-anchoring instances, extending spatial supervision from scene text to document-centric parsing.

\subsection{Continued Pretraining}
\label{sec:continued_cpt}

Building on the image-text spatial alignment established during text-anchoring pretraining, we further perform OCR-centric continued pretraining (CPT). Using a large-scale, multi-task OCR corpus, this stage expands task coverage and trains the model to produce outputs for multiple OCR tasks in a shared autoregressive format. It also provides the initialization for Mix-RL. We retain the vision-language backbone trained during the Q-Mask stage and remove the mask decoder. The model is trained with the standard next-token prediction objective; in addition, the single multi-token prediction (MTP) head already present in the base model is retained and jointly trained:
\begin{equation}
\mathcal{L}_{\mathrm{CPT}}(\theta)
=
-\mathbb{E}_{(I,c,y)\sim\mathcal{D}}
\left[
\sum_{t=1}^{|y|}
\log \pi_{\theta}\!\left(y_t \mid y_{<t}, I, c\right)
\right]
+
\lambda_{\mathrm{MTP}}\mathcal{L}_{\mathrm{MTP}}(\theta),
\end{equation}
where $\mathcal{L}_{\mathrm{MTP}}(\theta)$ denotes the auxiliary loss associated with the MTP head and $\lambda_{\mathrm{MTP}}$ is its weight.

The continued-pretraining corpus is centered on document parsing data produced by the data engine, covering recognition from region crops and direct parsing from full-page images (Section~\ref{sec:parsing_data_granularity}). It is complemented by OCR-centric understanding supervision, OCR-centric captioning, scene text, handwritten and calligraphic data, and long-tail tasks such as chemical formulas and charts.

\subsection{Reinforcement Learning}
\label{sec:rl}

Token-level cross-entropy in supervised training primarily constrains the model to imitate the target sequence token by token, and does not directly penalize document-level issues such as structural errors, missing fields, layout relations, or reading order~\cite{lu2026ovisocr2,jiang2026abotocr,chen2025logicsparsing}. We therefore apply Mix-RL after continued pretraining, jointly optimizing document parsing and OCR-centric understanding with task-specific rewards on mined hard examples.

We adopt a DAPO-style policy optimization objective~\citep{yu2026dapo}, building on GRPO~\citep{shao2024deepseekmath} with token-level loss aggregation, clip-higher, and dynamic sampling:
\begin{equation}
\mathcal{J}_{\mathrm{GRPO}}(\theta)
=\mathbb{E}_{x\sim\mathcal{D},\ \{o_i\}_{i=1}^{G}\sim\pi_{\mathrm{old}}}
\left[
\frac{1}{\sum_{i=1}^{G}|o_i|}
\sum_{i=1}^{G}\sum_{t=1}^{|o_i|}
\min\!\left(
 r_{i,t}(\theta)\hat{A}_i,
 \operatorname{clip}\!\left(r_{i,t}(\theta),1-\varepsilon_{\mathrm{low}},1+\varepsilon_{\mathrm{high}}\right)\hat{A}_i
\right)
\right],
\label{eq:grpo}
\end{equation}
where $r_{i,t}(\theta)$ is the token-level importance ratio, $\hat{A}_i$ is the group-normalized advantage, and $\varepsilon_{\mathrm{low}},\varepsilon_{\mathrm{high}}$ are the decoupled clipping bounds.

The objective in Eq.~\eqref{eq:grpo} does not include an explicit KL penalty to a reference policy. Within each rollout group, token-level aggregation assigns the same loss weight to each generated token rather than averaging sequence-normalized losses, which is relevant for OCR outputs with substantial length variation. Clip-higher relaxes the upper clipping bound to accommodate beneficial probability increases, and dynamic sampling discards rollout groups with zero within-group reward variance so that policy updates use groups with informative relative reward differences. We use this objective to optimize document-parsing and OCR-centric understanding rewards.

The RL data are obtained from hard cases mined by the data engine using multi-factor sample mining (Section~\ref{sec:sample_mining}). After secondary filtering and review by both a stronger language model and human annotators, the retained samples form a training pool covering document parsing (text, tables, formulas, and full-page documents) and OCR-centric understanding. During training, dynamic sampling filters rollout groups by their reward variance.

The reward function is routed by task type. For region-level document parsing samples, we use text edit similarity, table TEDS, or formula CDM, depending on the sample type. For full-page document parsing samples, multi-granularity adaptive matching (MGAM)~\cite{wang2026mineru25pro} first aligns predicted elements with the ground-truth elements. Similarities are then computed by element type and aggregated using the modality character coverage in the ground truth:
\begin{equation}
R_{\mathrm{doc}}
= w_{\mathrm{text}}\,s_{\mathrm{text}}
+ w_{\mathrm{table}}\,s_{\mathrm{table}}
+ w_{\mathrm{formula}}\,s_{\mathrm{formula}},
\end{equation}
where $s_{\mathrm{text}}$, $s_{\mathrm{table}}$, and $s_{\mathrm{formula}}$ denote text edit similarity, table TEDS, and formula CDM, respectively, and the character coverage of each modality in the ground truth determines the weights. Within OCR-centric understanding, rewards follow the subtask format. For KIE samples, the model output first undergoes JSON validity checking, followed by entity- and field-level matching; the reward is computed from the normalized edit similarity of field values, with missing fields receiving zero score. For OCR-VQA samples, we use the Average Normalized Levenshtein Similarity (ANLS)~\cite{Biten_2019_ICCV} to evaluate answer correctness. Outputs that cannot be parsed, violate the required format, or exceed the length limit receive a zero reward.

%% file: sections/evaluation.tex
\input{tables/omnidocbench_v16_tables}
\FloatBarrier
\section{Evaluation}
We evaluate content and structure recovery, robustness to acquisition artifacts and uncommon writing styles, and the ability to use document content for question answering and structured extraction. The benchmarks cover full-page parsing, individual elements, historical scripts, OCR-VQA, and KIE.

\subsection{Inference Setup and Metrics}
\label{sec:eval_setup}
\paragraph{Document-parsing inference.}
The main OmniDocBench v1.6, Real5, and Wild comparisons use two-stage inference: PP-DocLayoutV3~\citep{cui2026rtdoclayoutrealtimeendtoenddocument} detects regions, \METHODNAME recognizes each crop, and the predictions are assembled into a page-level output. The reported 0.8B parameter count excludes the external layout detector.

\paragraph{Metrics and aggregation.}
For document parsing, we report text and reading-order normalized edit distances (lower is better), formula CDM, and table TEDS and TEDS-S. TEDS measures both structure and cell content; TEDS-S measures structure alone. Each parsing benchmark retains its own Overall score.

For OCR-oriented question answering, Overall is the arithmetic mean of DocVQA, InfoVQA, ChartQA, OCRBench, and TextVQA, with all five scores expressed on a 0--100 scale. Models with missing results retain their available individual scores but are not assigned an Overall score. KIE is evaluated by normalized exact match after lenient JSON parsing and key-and-value normalization; task-level and document-category results are reported separately.

\subsection{Document Parsing}
\input{tables/real5omnidocbench}

\subsubsection{Standard Documents: OmniDocBench v1.6}
OmniDocBench v1.6~\citep{wang2026mineru25pro} jointly tests recovery of text, formulas, tables, and reading order across 1,651 pages with diverse layouts. Its adaptive matching reduces sensitivity to predicted element segmentation, supporting comparison across parsers with different output granularities.

With a 0.8B VLM, our two-stage system reaches 96.83 Overall, close to TeleOCR's 96.87 with a 1.2B VLM and above the same-size OvisOCR2 at 96.58 (Table~\ref{tab:omnidocbench_v16_external}). Formula recognition is its clearest strength: CDM reaches 98.50, the highest in the comparison. OvisOCR2 retains lower text and reading-order edit distances, and TeleOCR has stronger table scores. The formula advantage is consistent with the data engine's use of CDM-based expert agreement and render-guided correction of disputed labels (Section~\ref{sec:data_engine}). This suggests a task-specific bias: annotation choices may shape the model's relative strengths across parsing tasks.

\subsubsection{Acquisition Robustness: Real5-OmniDocBench and Wild-OmniDocBench}
Real5-OmniDocBench~\citep{real5omnidocbench} tests parsing under five acquisition conditions. It contains 6,775 recaptured images of OmniDocBench v1.5 pages, covering scanning, warping, screen photography, uneven illumination, and skew. Our system leads the compared methods both overall (95.24) and under each of the five conditions (Table~\ref{tab:real5omnidocbench}). Relative to PaddleOCR-VL-1.6, the Overall gain is 2.05 points, rising to 2.98 on warped pages and 2.11 under uneven illumination. Gains across all five conditions indicate robustness to both geometric deformation and appearance changes.

Wild-OmniDocBench~\citep{li2026towards} extends this test to documents recaptured in physical scenes, with glare, moir\'e patterns, perspective distortion, blur, and uneven illumination. At 87.94 Overall, our system trails TeleOCR's 88.53 while attaining the lowest reading-order edit distance in the comparison, 0.1931 (Table~\ref{tab:wildomnidocbench}). Recovering element order remains a strength under difficult capture conditions; table reconstruction leaves more room for improvement relative to TeleOCR.
\input{tables/wildomnidocbench}

\subsubsection{Element-Level Recognition}
\begin{table}[b]
\centering
\begin{minipage}[t]{0.36\textwidth}
\input{tables/pubtabnet}
\end{minipage}\hfill
\begin{minipage}[t]{0.63\textwidth}
\input{tables/unimer}
\end{minipage}
\end{table}

Element-level tests isolate recognition from full-page layout recovery. PubTabNet~\citep{pubtabnet} measures reconstruction of table structure and cell content through TEDS. Our score of 92.27 exceeds all methods listed in Table~\ref{tab:pubtabnet}, supporting its ability to reconstruct individual tables. UniMER-Test~\citep{unimernet} probes formula recognition across four subsets, CPE, HWE, SCE, and SPE. We obtain the highest Overall score in Table~\ref{tab:unimer}, 97.84, and the highest SPE score, 99.460. Together with OmniDocBench, this provides evidence of strong formula recognition across both page-level and element-level evaluations.

\subsubsection{Long-Tail Recognition: Historical Scripts}
Historical handwriting and calligraphy probe a parser's recognition limits beyond modern printed documents. We use the 1,600-image mature-script subset of Chronicles-OCR~\citep{li2026chronicles}, with 400 images each of Clerical, Regular, Running, and Cursive scripts. \METHODNAME reaches 0.73 on Mature Average, exceeding Seed2.0 Pro's 0.72 (Table~\ref{tab:chronicles_ocr}). This extends the recognition component of parsing to long-tail writing styles, consistent with our coverage-driven synthesis of calligraphy and uncommon glyph combinations.

\subsection{OCR-Centric Understanding}
\label{sec:ocr_vqa_evaluation}
\subsubsection{OCR-Oriented Visual Question Answering}
\input{tables/ocr_vqa}

\input{tables/Chronicles-OCR}
OCR-VQA tests whether the model can locate relevant text and interpret it in context. DocVQA~\citep{mathew2021docvqa} tests question answering over documents; InfoVQA~\citep{mathew2021infographicvqa} and ChartQA~\citep{masry2022chartqa} extend the setting to infographics and charts; TextVQA~\citep{singh2019towards} covers text in natural images. OCRBench~\citep{liu2024ocrbench} broadens the assessment with both recognition and question-answering tasks.

The five-benchmark mean is 83.2, a gain of 10.3 points over Qwen3.5-0.8B and 6.7 over the similarly sized HunyuanOCR-1.5 (Table~\ref{tab:ocr_vqa}). Relative to Qwen3.5-0.8B, gains occur on all five benchmarks and are largest on infographics (InfoVQA, +14.8) and charts (ChartQA, +15.1). Our 0.8B model also exceeds Qwen3.5-2B by 2.3 points overall and approaches the general-purpose Qwen3.5-4B, trailing it by 1.7 points while using one-fifth as many parameters.

\subsubsection{Key Information Extraction}
\paragraph{Benchmark and protocol.}
KIE tests selection of requested fields and generation of structured records in a prescribed schema. Our in-house benchmark contains 1,655 document photos and app screenshots covering document-field, flight-information, and address extraction. Document categories include ID cards, bank cards, passports, driver's licenses, invoices, and ID-card images embedded in chat screenshots. Flight samples require flight numbers and departure and arrival cities; address samples cover delivery addresses, meeting locations, and restaurant addresses.

\paragraph{Results.}
At 59.09\% overall accuracy, \METHODNAME exceeds Qwen3.5-2B by 7.49 percentage points (Table~\ref{tab:kie_task}). The gain is concentrated in flight-information extraction (73.00\% vs.\ 43.00\%) and document-field extraction (77.89\% vs.\ 63.15\%), showing stronger recovery of named fields into structured records. Address extraction remains difficult for both models, at 45.93\% and 45.30\%. The document-category breakdown further localizes the limitation: we lead on five of six categories, but trail Qwen3.5-2B on invoices (65.00\% vs.\ 77.00\%; Table~\ref{tab:kie_card}). Address and invoice extraction remain clear targets for further data development.
\input{tables/kie_card}
\newpage

%% file: tables/omnidocbench_v16_tables.tex
%

\begin{table}[t]
\centering
\caption{Comparison on OmniDocBench v1.6~\citep{wang2026mineru25pro}.}
\label{tab:omnidocbench_v16_external}
\footnotesize
\setlength{\tabcolsep}{3.8pt}
\renewcommand{\arraystretch}{1.08}
\resizebox{\textwidth}{!}{%
\begin{tabular}{@{}>{\raggedright\arraybackslash}p{4.25cm}c*{6}{c}@{}}
\toprule
\textbf{Method} & \textbf{Size} & \textbf{Overall}$\uparrow$ & \textbf{Text Edit}$\downarrow$ & \textbf{Formula CDM}$\uparrow$ & \textbf{Table TEDS}$\uparrow$ & \textbf{Table TEDS-S}$\uparrow$ & \textbf{Reading Order Edit}$\downarrow$ \\
\midrule

\rowcolor{gray!12}
\multicolumn{8}{@{}l}{\textit{General VLMs}}\\
Ovis2.6-30B-A3B~\citep{lu2025ovis25} & 30B  & 93.70 & 0.035 & 95.17 & 89.44 & 92.40 & 0.135 \\
Gemini 3 Pro~\citep{google2025new} & -   & 92.91 & 0.064 & 95.99 & 89.15 & 92.96 & 0.165 \\
Gemini 3 Flash~\citep{doshi2025gemini} & -   & 92.62 & 0.066 & 95.16 & 89.29 & 93.51 & 0.172 \\
Qwen3-VL-235B~\citep{bai2025qwen3vl} & 235B & 89.78 & 0.063 & 92.55 & 83.07 & 86.75 & 0.166 \\
GPT-5.2~\citep{openai2025introducing} & -   & 86.59 & 0.114 & 88.21 & 82.95 & 87.93 & 0.193 \\
Kimi K2.5~\citep{kimi2026kimik25} & 1T   & 84.53 & 0.107 & 83.50 & 80.76 & 84.00 & 0.211 \\
InternVL3.5-241B~\citep{wang2025internvl35} & 241B & 83.76 & 0.130 & 89.95 & 74.35 & 79.78 & 0.215 \\
\addlinespace[2pt]

\rowcolor{gray!12}
\multicolumn{8}{@{}l}{\textit{Specialized OCR Models}}\\
Marker~\citep{datalab2025marker} & --   & 78.44 & 0.157 & 85.24 & 65.77 & 73.24 & 0.243 \\
POINTS-Reader~\citep{liu2025pointsreader} & 3B         & 83.37 & 0.096 & 85.72 & 73.98 & 77.40 & 0.198 \\
Nanonets-OCR-s~\citep{nanonets2025nanonetsocrs} & 3B         & 83.61 & 0.108 & 81.46 & 80.18 & 84.51 & 0.213 \\
Mistral OCR~\citep{mistral2025mistral} & -         & 85.66 & 0.097 & 89.91 & 76.78 & 80.93 & 0.171 \\
olmOCR~\citep{poznanski2025olmocr} & 7B         & 85.74 & 0.139 & 88.10 & 83.00 & 87.17 & 0.216 \\
Dolphin-1.5~\citep{feng2025dolphin} & 0.3B & 86.52 & 0.094 & 87.49 & 81.43 & 84.82 & 0.167 \\
MonkeyOCR-pro-3B~\citep{zhang2026monkeyocr} & 3B   & 88.57 & 0.074 & 88.74 & 84.35 & 88.62 & 0.189 \\
OCRVerse~\citep{zhong2026ocrverse} & 4B         & 88.60 & 0.063 & 89.61 & 82.44 & 86.27 & 0.163 \\
Dolphin-v2~\citep{feng2026dolphinv2} & 3B   & 89.50 & 0.069 & 91.01 & 84.40 & 87.44 & 0.150 \\
DeepSeek-OCR 2~\citep{wei2026deepseekocr} & 3B         & 90.25 & 0.050 & 91.84 & 83.89 & 87.75 & 0.144 \\
OpenDoc-0.1B~\citep{du2025unirec01b} & 0.1B       & 90.67 & 0.049 & 93.02 & 83.88 & 87.45 & 0.140 \\
dots.ocr~\citep{li2025dotsocr} & 3B         & 90.77 & 0.048 & 89.95 & 87.18 & 90.58 & 0.138 \\
FireRed-OCR~\citep{wu2026fireredocr} & 2B         & 93.26 & 0.037 & 95.44 & 88.04 & 91.06 & 0.131 \\
ABot-OCR~\citep{jiang2026abotocr} & 2B         & 93.30 & 0.037 & 94.86 & 88.69 & 91.87 & 0.137 \\
Logics-Parsing-v2~\citep{chen2025logicsparsing} & 4B         & 93.33 & 0.041 & 95.65 & 88.42 & 91.98 & 0.137 \\
Youtu-Parsing~\citep{cao2026youtuparsing} & 2.5B & 93.74 & 0.044 & 93.63 & 92.02 & 95.00 & 0.116 \\
Qianfan-OCR~\citep{dong2026qianfanocr} & 4B         & 93.90 & 0.040 & 95.08 & 90.53 & 93.31 & 0.130 \\
Unlimited-OCR~\citep{yin2026unlimited} & 3B-A0.5B   & 93.92 & 0.042 & 95.79 & 90.16 & 93.32 & 0.129 \\
PaddleOCR-VL~\citep{cui2025paddleocrvl} & 0.9B & 94.18 & 0.040 & 95.91 & 90.65 & 93.74 & 0.135 \\
HunyuanOCR-1.5~\citep{li2026hunyuanocr15} & 1B & 94.74 & 0.039 & 94.50 & 93.67 & 94.71 & 0.129 \\
PaddleOCR-VL-1.5~\citep{cui2026paddleocrvl15} & 0.9B & 94.93 & 0.038 & 96.89 & 91.67 & 94.37 & 0.130 \\
GLM-OCR~\citep{duan2026glmocr} & 0.9B & 95.22 & 0.044 & 97.18 & 92.83 & 95.39 & 0.133 \\
MinerU2.5-Pro~\citep{wang2026mineru25pro} & 1.2B & 95.75 & 0.036 & 97.45 & 93.42 & 95.92 & 0.120 \\
PaddleOCR-VL-1.6~\citep{zhang2026paddleocrvl16} & 0.9B & 96.33 & 0.033 & 97.49 & 94.76 & 97.11 & 0.127 \\
OvisOCR2~\citep{lu2026ovisocr2} & 0.8B       & 96.58 & \textbf{0.025} & 97.53 & 94.76 & 97.16 & \textbf{0.111} \\
TeleOCR~\citep{cai2026navidc} & 1.2B & \textbf{96.87} & 0.027 & 96.36 & \textbf{97.05} & \textbf{98.52} & 0.122 \\
\rowcolor{Orange}
Ours          & 0.8B & 96.83 & 0.031 & \textbf{98.50} & 95.11 & 97.19 & 0.122 \\
\bottomrule
\end{tabular}%
}
\vspace{2pt}
\end{table}

%% file: tables/real5omnidocbench.tex
\begin{table}[tbp]
\centering
\caption{Comparison on Real5-OmniDocBench \cite{real5omnidocbench}.}
\label{tab:real5omnidocbench}
\footnotesize
\setlength{\tabcolsep}{4pt}
\resizebox{\textwidth}{!}{%
\begin{tabular}{llcccccc}
\toprule
Method & Size & Overall$\uparrow$ & Scanning$\uparrow$ & Warping$\uparrow$ & Screen-Photo$\uparrow$ & Illumination$\uparrow$ & Skew$\uparrow$ \\
\midrule
\rowcolor{gray!12}
\multicolumn{8}{@{}l}{\textit{General VLMs}}\\
Kimi-K2.6~\citep{moonshot2026kimik26} & 1T & 89.76 & 90.08 & 89.62 & 89.58 & 89.91 & 89.61 \\
Gemini 3 Pro \cite{google2025new} & - & 89.24 & 89.47 & 88.90 & 88.86 & 89.53 & 89.45 \\
Kimi K2.5 \cite{kimi2026kimik25} & 1T & 89.09 & 89.67 & 88.86 & 88.39 & 89.66 & 88.86 \\
Doubao-Seed-2.1-Pro~\citep{bytedanceseed2026doubaoseed21pro} & - & 89.02 & 88.85 & 89.36 & 88.99 & 89.13 & 88.79 \\
Qwen3-VL-235B \cite{bai2025qwen3vl} & 235B & 88.90 & 89.43 & 89.99 & 89.27 & 89.27 & 86.56 \\
Gemini 2.5 Pro \cite{gemini25pro} & - & 88.21 & 89.25 & 87.63 & 87.11 & 87.97 & 89.07 \\
Qwen2.5-VL-72B~\citep{bai2025qwen25vl} & 72B & 86.92 & 86.19 & 87.77 & 86.48 & 87.25 & 86.90 \\
GPT-5.2 \cite{openai2025introducing} & - & 78.66 & 84.43 & 76.26 & 76.75 & 80.88 & 75.00 \\
\addlinespace[2pt]
\rowcolor{gray!12}
\multicolumn{8}{@{}l}{\textit{Specialized OCR Models}}\\
Marker-1.8.2 \cite{datalab2025marker} & -- & 60.10 & 70.27 & 58.98 & 63.65 & 66.31 & 41.27 \\
Dolphin-1.5~\citep{feng2025dolphin} & 0.3B & 61.48 & 83.39 & 50.50 & 69.76 & 75.61 & 28.16 \\
Dolphin \cite{feng2025dolphin} & 322M & 61.78 & 72.16 & 60.35 & 64.29 & 67.29 & 44.83 \\
PP-StructureV3~\citep{cui2025paddleocr3} & -- & 64.45 & 84.68 & 59.34 & 66.89 & 73.38 & 37.98 \\
DeepSeek-OCR 2 \cite{wei2026deepseekocr} & 3B & 73.01 & 89.59 & 66.53 & 71.65 & 76.02 & 61.28 \\
DeepSeek-OCR \cite{wei2025deepseekocr} & 3B & 73.99 & 86.17 & 67.20 & 75.31 & 78.10 & 63.01 \\
MinerU2-VLM~\citep{opendatalab2025mineru2vlm} & 0.9B & 76.95 & 83.60 & 73.73 & 78.77 & 80.51 & 68.16 \\
MonkeyOCR-pro-1.2B \cite{zhang2026monkeyocr} & 1.9B & 77.15 & 84.64 & 76.59 & 80.24 & 82.11 & 62.18 \\
MonkeyOCR-3B \cite{zhang2026monkeyocr} & 3.7B & 78.29 & 84.65 & 77.27 & 80.71 & 83.16 & 65.67 \\
MonkeyOCR-pro-3B \cite{zhang2026monkeyocr} & 3.7B & 79.49 & 86.94 & 78.90 & 82.44 & 84.71 & 64.47 \\
Nanonets-OCR-s \cite{nanonets2025nanonetsocrs} & 3B & 84.19 & 85.52 & 83.56 & 84.86 & 85.01 & 81.98 \\
PaddleOCR-VL \cite{cui2025paddleocrvl} & 0.9B & 85.54 & 92.11 & 85.97 & 82.54 & 89.61 & 77.47 \\
MinerU2.5 \cite{niu2025mineru25} & 1.2B & 85.61 & 90.06 & 83.76 & 89.41 & 89.57 & 75.24 \\
dots.ocr \cite{li2025dotsocr} & 3B & 86.38 & 86.87 & 86.01 & 87.18 & 87.57 & 84.27 \\
MonkeyOCRv2-S-Parsing~\citep{liu2026monkeyocrv2} & 0.6B & 87.90 & 88.87 & 88.17 & 87.64 & 86.75 & 88.09 \\
MinerU2.5-Pro \cite{wang2026mineru25pro} & 1.2B & 88.94 & 92.11 & 88.72 & 91.29 & 91.31 & 81.26 \\
MonkeyOCRv2-B-Parsing~\citep{liu2026monkeyocrv2} & 0.7B & 89.22 & 89.49 & 89.70 & 88.40 & 88.54 & 89.97 \\
GLM-OCR \cite{duan2026glmocr} & 0.9B & 90.32 & 92.67 & 90.68 & 91.75 & 91.12 & 85.39 \\
PaddleOCR-VL-1.5 \cite{cui2026paddleocrvl15} & 0.9B & 92.05 & 93.43 & 91.25 & 91.76 & 92.16 & 91.66 \\
OvisOCR2 \cite{lu2026ovisocr2} & 0.8B & 92.29 & 93.77 & 91.40 & 93.09 & 92.88 & 90.33 \\
PaddleOCR-VL-1.6 \cite{zhang2026paddleocrvl16} & 0.9B & 93.19 & 94.74 & 92.48 & 92.78 & 93.28 & 92.66 \\
\rowcolor{Orange}
Ours & 0.8B & \textbf{95.24} & \textbf{96.41} & \textbf{95.46} & \textbf{94.47} & \textbf{95.39} & \textbf{94.46} \\
\bottomrule
\end{tabular}%
}
\end{table}

%% file: tables/wildomnidocbench.tex
\begin{table}[tbp]
\centering
\caption{Comparison on Wild-OmniDocBench \cite{li2026towards}.}
\label{tab:wildomnidocbench}
\footnotesize
\setlength{\tabcolsep}{4pt}
\resizebox{\textwidth}{!}{%
\begin{tabular}{llcccccc}
\toprule
\textbf{Method} & \textbf{Size} & \textbf{Overall}$\uparrow$ & \textbf{Text Edit}$\downarrow$ & \textbf{Formula CDM}$\uparrow$ & \textbf{Table TEDS}$\uparrow$ & \textbf{Table TEDS-S}$\uparrow$ & \textbf{Reading Order Edit}$\downarrow$ \\
\midrule
\rowcolor{gray!12}
\multicolumn{8}{@{}l}{\textit{Specialized OCR Models}}\\
Logics-Parsing-v2 \cite{chen2025logicsparsing} & 4B & 77.10 & 0.4029 & \textbf{91.40} & 80.19 & 87.16 & 0.2355 \\
HunyuanOCR-1.5 \cite{li2026hunyuanocr15} & 1B & 77.62 & 0.1979 & 85.12 & 67.54 & 70.67 & 0.2750 \\
dots.ocr \cite{li2025dotsocr} & 3B & 81.84 & 0.1483 & 85.00 & 75.32 & 80.20 & 0.2200 \\
PaddleOCR-VL-1.5 \cite{cui2026paddleocrvl15} & 0.9B & 84.64 & 0.1461 & 86.72 & 81.80 & 86.52 & 0.2138 \\
GLM-OCR \cite{duan2026glmocr} & 0.9B & 85.08 & 0.1514 & 89.09 & 81.31 & 85.90 & 0.2228 \\
MinerU2.5-Pro \cite{wang2026mineru25pro} & 1.2B & 87.33 & 0.1362 & 90.15 & 85.46 & 90.12 & 0.2013 \\
PaddleOCR-VL-1.6 \cite{zhang2026paddleocrvl16} & 0.9B & 87.36 & 0.1369 & 88.42 & 85.76 & 90.14 & 0.2057 \\
OvisOCR2 \cite{lu2026ovisocr2} & 0.8B & 87.91 & 0.1290 & 90.37 & 85.13 & 89.11 & 0.2021 \\
TeleOCR~\citep{cai2026navidc} & 1.2B & \textbf{88.53} & \textbf{0.1173} & 88.26 & \textbf{89.05} & \textbf{92.14} & 0.2011 \\
\rowcolor{Orange}
Ours & 0.8B & 87.94 & 0.1233 & 89.55 & 86.61 & 90.94 & \textbf{0.1931} \\
\bottomrule
\end{tabular}%
}
\end{table}

%% file: tables/pubtabnet.tex
\captionof{table}{Comparison on PubTabNet~\citep{pubtabnet}, measured by TEDS.}
\label{tab:pubtabnet}
\centering
\footnotesize
\setlength{\tabcolsep}{4pt}
\scalebox{0.85}{%
\begin{tabular}{lcc}
\toprule
Model & Size & TEDS$\uparrow$ \\
\midrule
LightOnOCR-2~\citep{taghadouini2026lightonocr} & 1B & 51.42 \\
Qwen3.5-2B~\citep{qwen2026qwen35} & 2B & 76.33 \\
Qwen2.5-VL-7B~\citep{bai2025qwen25vl} & 7B & 81.60 \\
olmOCR-2~\citep{poznanski2025olmocr} & 7B & 84.02 \\
GPT-5.2~\citep{openai2025introducing} & - & 84.40 \\
PaddleOCR-VL-1.5~\citep{cui2026paddleocrvl15} & 0.9B & 84.60 \\
GLM-OCR~\citep{duan2026glmocr} & 0.9B & 85.20 \\
MinerU2.5~\citep{niu2025mineru25} & 1.2B & 89.07 \\
DeepSeek-OCR 2~\citep{wei2026deepseekocr} & 3B & 89.53 \\
Qwen3.5-35B-A3B~\citep{qwen2026qwen35} & 35B-A3B & 90.06 \\
Gemini 3 Pro~\citep{google2025new} & -     & 91.40 \\
Infinity-Parser-7B~\citep{wang2025infinityparser} & 7B & 91.82 \\
\rowcolor{Orange}
Ours & 0.8B & \textbf{92.27} \\
\bottomrule
\end{tabular}
}
\vspace{1mm}

%% file: tables/unimer.tex
\captionof{table}{Comparison on the UniMER-Test benchmark~\citep{unimernet}, which contains 23{,}757 real-world mathematical expressions.} 
\label{tab:unimer}
\centering
\footnotesize
\setlength{\tabcolsep}{3.5pt}
\begin{tabular}{lccccc}
\toprule
Model & CPE$\uparrow$ & HWE$\uparrow$ & SCE$\uparrow$ & SPE$\uparrow$ & Overall$\uparrow$ \\
\midrule
DeepSeek-OCR 2~\citep{wei2026deepseekocr} & 91.97 & 81.67 & 77.19 & 95.51 & 86.59 \\
FireRed-OCR~\citep{wu2026fireredocr} & 94.35 & 85.42 & 89.94 & 96.75 & 91.62 \\
PaddleOCR-VL-1.5~\citep{cui2026paddleocrvl15} & 98.84 & 92.27 & 94.95 & 99.27 & 96.33 \\
Qwen3-VL-235B~\citep{bai2025qwen3vl} & 97.47 & 94.23 & 96.21 & 98.46 & 96.59 \\
MinerU2.5~\citep{niu2025mineru25} & 97.79 & 94.42 & 96.65 & 98.57 & 96.86 \\
GLM-OCR~\citep{duan2026glmocr} & 96.74 & 95.10 & \textbf{97.77} & 98.42 & 97.01 \\
Infinity-Parser2-Pro~\citep{infinityparser2} & 98.30 & 96.70 & 96.20 & 99.40 & 97.70 \\
MinerU2.5-Pro~\citep{wang2026mineru25pro} & \textbf{98.97} & 95.38 & 97.04 & 99.44 & 97.71 \\
Qwen3.5-A17B~\citep{qwen2026qwen35} & 98.32 & \textbf{97.59} & 95.87 & 99.41 & 97.80 \\
\rowcolor{Orange}
Ours & 98.55 & 95.82 & 97.51 & \textbf{99.46} & \textbf{97.84} \\
\bottomrule
\end{tabular}

%% file: tables/ocr_vqa.tex
\begin{table}[tbp]
\centering
\caption{Comparison on document-oriented visual question answering benchmarks. Overall is the arithmetic mean of DocVQA, InfoVQA, ChartQA, OCRBench, and TextVQA on a 0--100 scale; it is reported only when all five scores are available. ``--'' denotes unavailable results.}
\label{tab:ocr_vqa}
\footnotesize
\setlength{\tabcolsep}{4.6pt}
\renewcommand{\arraystretch}{1.08}
\begin{tabular}{@{}l c c c c c c c@{}}
\toprule
\textbf{Model} & \textbf{Size} & \textbf{DocVQA}$\uparrow$ & \textbf{InfoVQA}$\uparrow$ & \textbf{ChartQA}$\uparrow$ & \textbf{OCRBench}$\uparrow$ & \textbf{TextVQA}$\uparrow$ & \textbf{Overall}$\uparrow$ \\
\midrule
\rowcolor{gray!12}
\multicolumn{8}{@{}l}{\textit{General VLMs}}\\
GPT-5.2~\citep{openai2025introducing} & - & 91.7 & 84.0 & 57.0 & 80.7 & 72.8 & 77.2 \\
GLM-4.5V~\citep{glm45v} & 106B-A12B & 94.5 & 84.1 & 86.6 & 87.2 & 72.0 & 84.9 \\
Gemini 3 Pro~\citep{google2025new} & - & -- & -- & 57.2 & 94.0 & -- & -- \\
Qwen3.5-0.8B~\citep{qwen2026qwen35} & 0.8B & 88.5 & 60.3 & 69.5 & 77.9 & 68.3 & 72.9 \\
Qwen3.5-2B~\citep{qwen2026qwen35} & 2B & 92.4 & 72.4 & 77.0 & 85.9 & 76.9 & 80.9 \\
Qwen3.5-4B~\citep{qwen2026qwen35} & 4B & 94.4 & 80.4 & 82.4 & 86.6 & 80.8 & \textbf{84.9} \\
Gemma-4-E2B-it~\citep{gemmateam2026gemma4} & 2B & 73.8 & 38.1 & 42.6 & 72.4 & 59.7 & 57.3 \\
Gemma-4-E4B-it~\citep{gemmateam2026gemma4} & 4B & 79.2 & 47.3 & 38.4 & 76.0 & 66.1 & 61.4 \\
MiniCPM-V-4.5~\citep{minicpmv45} & 8B & 84.9 & 69.6 & 87.4 & 89.0 & 82.2 & 82.6 \\
\addlinespace[2pt]
\rowcolor{gray!12}
\multicolumn{8}{@{}l}{\textit{Specialized OCR Models}}\\
HunyuanOCR~\citep{hunyuan2025hunyuanocr} & 1B & 86.8 & 61.6 & 78.5 & 86.0 & 71.1 & 76.8 \\
HunyuanOCR-1.5~\citep{li2026hunyuanocr15} & 1B & 87.6 & 55.2 & 78.3 & 86.1 & 75.1 & 76.5 \\
TokenVL-2B~\citep{TokenFD} & 2B & 89.9 & 61.0 & 81.1 & 82.1 & 76.4 & 78.1 \\
Mini-Monkey~\citep{MiNi-Monkey} & 2B & 87.4 & 60.1 & 76.5 & -- & -- & -- \\
TextHawk2~\citep{yu2024texthawk2} & 7B & 89.6 & 67.8 & 81.4 & 78.4 & 75.1 & 78.5 \\
MonkeyOCRv2-S-Und~\citep{liu2026monkeyocrv2} & 1.7B & 79.3 & 44.5 & 62.0 & 52.2 & -- & -- \\
MonkeyOCRv2-B-Und~\citep{liu2026monkeyocrv2} & 1.8B & 79.3 & 46.3 & 62.0 & 58.1 & -- & -- \\
\addlinespace[2pt]
\rowcolor{Orange}
\textbf{Ours} & 0.8B & 93.1 & 75.1 & 84.6 & 84.6 & 78.6 & \textbf{83.2} \\
\bottomrule
\end{tabular}
\end{table}

%% file: tables/Chronicles-OCR.tex
\begin{table}[htbp]
\centering
\caption{Comparison on mature-script (Clerical,
Regular, Running, and Cursive) subset of Chronicles-OCR~\cite{li2026chronicles}.}
\label{tab:chronicles_ocr}
\footnotesize
\renewcommand{\arraystretch}{1.0}
\setlength{\tabcolsep}{4pt}
\begin{tabular}{llcl}
\toprule
Model Type & Model & Model Size & Average \\
\midrule
\multirow{11}{*}{\shortstack{Open-source\\ General VLMs}}
 & Gemma 4 31B it \cite{gemmateam2026gemma4} & 31B & 0.35 \\
 & InternVL3.5-8B \cite{wang2025internvl35} & 8B & 0.39 \\
 & MiniCPM-V 4.5 \cite{minicpmv45} & 8B & 0.40 \\
 & GLM-4.5V \cite{glm45v} & 106B-A12B & 0.43 \\
 & Ovis2.6-30B-A3B \cite{lu2025ovis25} & 30B-A3B & 0.51 \\
 & InternVL3.5-A28B \cite{wang2025internvl35} & 241B-A28B & 0.56 \\
 & Qwen3.5-9B \cite{qwen2026qwen35} & 9B & 0.60 \\
 & Qwen3-VL-8B \cite{bai2025qwen3vl} & 8B & 0.65 \\
 & Qwen3-VL-A22B \cite{bai2025qwen3vl} & 235B-A22B & 0.66 \\
 & Kimi K2.5 \cite{kimi2026kimik25} & 1T & 0.71 \\
 & Qwen3.5-A17B \cite{qwen2026qwen35} & 397B-A17B & 0.73 \\
\midrule
\multirow{7}{*}{\shortstack{Proprietary\\ General VLMs}}
 & GPT-5 \cite{gpt5} & - & 0.41 \\
 & Claude Opus 4.7 \cite{claudeopus47} & - & 0.50 \\
 & Gemini 2.5 Pro \cite{gemini25pro} & - & 0.52 \\
 & MiMo-V2-Omni \cite{mimov2omni} & - & 0.55 \\
 & Seed1.8 \cite{seed18} & - & 0.67 \\
 & Gemini 3.1 Pro \cite{gemini31pro} & - & 0.68 \\
 & Seed2.0 Pro \cite{seed20pro} & - & 0.72 \\
\midrule
\multirow{7}{*}{\shortstack{Expert\\ OCR Models}}
 & Unlimited-OCR \cite{yin2026unlimited} & 3B-A0.5B & 0.21 \\
 & DeepSeek-OCR \cite{wei2025deepseekocr} & 3B-A0.5B & 0.24 \\
 & GLM-OCR \cite{duan2026glmocr} & 0.9B & 0.38 \\
 & PaddleOCR-VL-1.6 \cite{zhang2026paddleocrvl16} & 0.9B & 0.41 \\
 & dots.ocr \cite{li2025dotsocr} & 3B & 0.47 \\
 & HunyuanOCR-1.5 \cite{li2026hunyuanocr15} & 1B & \textbf{0.79} \\
\rowcolor{Orange}
 & \textbf{Ours} & 0.8B & 0.73 \\
\bottomrule
\end{tabular}
\end{table}

%% file: tables/kie_card.tex
\begin{table}[!htbp]
\centering
\caption{Task-level key information extraction accuracy (\%) on our in-house benchmark.}
\label{tab:kie_task}
\footnotesize
\setlength{\tabcolsep}{4pt}
\begin{tabular}{lccccc}
\toprule
 & \textbf{Ours} & \textbf{Qwen3.5-0.8B}~\citep{qwen2026qwen35} & \textbf{Qwen3.5-2B}~\citep{qwen2026qwen35} & \textbf{GLM-OCR}~\citep{duan2026glmocr} & \textbf{dots.ocr}~\citep{li2025dotsocr} \\
\midrule
Overall & \textbf{59.09} & 34.98 & 51.60 & 37.40 & 27.85 \\
Screen address & \textbf{45.93} & 42.38 & 45.30 & 30.48 & 32.99 \\
Flight & \textbf{73.00} & 20.00 & 43.00 & 0.00 & 0.00 \\
Card & \textbf{77.89} & 25.63 & 63.15 & 54.77 & 24.29 \\
\bottomrule
\end{tabular}
\vspace{4pt}
\captionof{table}{Structured extraction accuracy (\%) by document category on our in-house benchmark.}
\label{tab:kie_card}
\footnotesize
\setlength{\tabcolsep}{4pt}
\begin{tabular}{lccccc}
\toprule
Card type & \textbf{Ours} & \textbf{Qwen3.5-0.8B}~\citep{qwen2026qwen35} & \textbf{Qwen3.5-2B}~\citep{qwen2026qwen35} & \textbf{GLM-OCR}~\citep{duan2026glmocr} & \textbf{dots.ocr}~\citep{li2025dotsocr} \\
\midrule
ID card & \textbf{84.16} & 1.00 & 58.42 & 59.41 & 10.89 \\
Bank card & \textbf{79.80} & 11.11 & 76.77 & 58.59 & 30.30 \\
Passport & \textbf{81.00} & 22.00 & 73.00 & 66.00 & 1.00 \\
Driver's license & \textbf{79.59} & 57.14 & 59.18 & 75.51 & 34.69 \\
Chat-record ID & \textbf{77.79} & 0.00 & 34.34 & 0.00 & 2.02 \\
Invoice & 65.00 & 63.00 & \textbf{77.00} & 69.00 & 67.00 \\
\bottomrule
\end{tabular}
\end{table}

%% file: sections/analysis.tex
\FloatBarrier
\section{Analysis}

We investigate when understanding supervision begins to benefit parsing, how Q-Mask and Mix-RL improve the model, how model capacity affects the two capabilities, and how MOPD compares with Mix-RL over training.

\subsection{Emergent Transfer from Understanding to Parsing}
\label{sec:data_level_analysis}

We examine understanding-to-parsing transfer at early, middle, and late stages of parsing training (Table~\ref{tab:understanding_transfer_scale}). Region-level data supervise recognition from individual crops, while page-level data supervise parsing from complete images (Section~\ref{sec:parsing_data_granularity}). Within each ablation, we vary the training data and keep the remaining setup fixed, evaluating on OmniDocBench v1.6.

\paragraph{Early stage.}
The early-stage setting uses 9\% of all parsing data and includes only region-level supervision. Adding OCR-centric understanding data lowers Overall from 93.594 to 93.264 ($-0.330$), with worse results on every component metric. This suggests that the model needs a stronger parsing foundation before it can benefit from additional understanding supervision.

\paragraph{Middle stage.}
The middle-stage setting uses 29\% of all parsing data, with approximately one page-level sample for every five region-level samples. Adding page-level supervision raises Overall from 94.730 to 94.941 ($+0.211$), improving all six reported metrics. Adding the same understanding pool used in the early-stage experiment then raises Overall to 95.580 ($+0.639$). All component metrics improve again, with the largest gain in Table TEDS ($+1.024$). Page-level supervision strengthens the parsing foundation, and understanding supervision provides a further benefit.
\input{tables/analysis_transfer_scale}

\paragraph{Late stage.}
The late-stage ablation uses 50\% of all parsing data, maintaining the page-to-region sample ratio used at the middle stage. Page-level supervision continues to improve overall performance, and adding understanding data yields further gains in formula recognition, table reconstruction, and reading order. Positive transfer thus persists as parsing training progresses.

\begin{takeaway}
{\sffamily\bfseries\textcolor{xiaomiorange}{Takeaway 1.}} Understanding-to-parsing transfer emerges as the parsing foundation develops: understanding supervision hurts at the early stage but benefits parsing at the middle and late stages. Page-level supervision also improves the parsing foundation.
\end{takeaway}

\subsection{Effects of Q-Mask and Mix-RL}
\label{sec:training_analysis}

\label{sec:stage_wise_results}

Table~\ref{tab:stage_wise_omnidocbench} reports two component ablations, with the 0.8B baseline (72.21 Overall) included for reference. At a checkpoint trained on approximately 78\% of the full dataset, Q-Mask raises Overall from 96.0241 to 96.3198 ($+0.2957$), with the largest increase in Formula CDM ($+0.4931$). In the later ablation, Mix-RL raises Overall from 96.4739 at its starting checkpoint to 96.8277 ($+0.3538$), with larger gains in Table TEDS ($+0.7114$) and TEDS-S ($+0.5094$). These results suggest complementary roles: Q-Mask improves recognition during pretraining, while Mix-RL further improves table reconstruction. The two comparisons are evaluated at different training stages.

\input{tables/analysis_qmask_mixrl}

\subsection{Model Capacity: Parsing and Understanding}
\label{sec:ablation_rl}

We compare the task-specific 4B teachers with the 0.8B Mix-RL model to examine how model capacity affects parsing and understanding. For each task, the teacher and student use the same amount of task-specific training data: each 4B teacher uses only its domain subset, while Mix-RL trains on the combined parsing and understanding pools (Section~\ref{sec:rl}). Table~\ref{tab:posttraining_comparison} also includes the CPT model and the MOPD student; the next subsection compares MOPD and Mix-RL over training.

\paragraph{Teachers and evaluation.}

The 4B teachers start from Qwen3.5-4B-base and follow the same Q-Mask text-anchoring and OCR-centric continued-pretraining recipe as the 0.8B model. During post-training, the parsing teacher uses only the parsing subset and the VQA teacher uses only the understanding subset.

Table~\ref{tab:posttraining_comparison} compares the task-specific teachers, the CPT student, and the two post-trained students in two panels. Parsing uses the same six OmniDocBench v1.6 metrics as the main evaluation. Understanding uses the arithmetic mean of DocVQA, InfoVQA, ChartQA, OCRBench, and TextVQA, matching the evaluation protocol and training curves. The Mix-RL rows report the same model results as the Ours rows in Tables~\ref{tab:omnidocbench_v16_external} and~\ref{tab:ocr_vqa}.

\input{tables/analysis_rl_omnidocbench}

\paragraph{Understanding places greater demands on model capacity.}

Scaling from 0.8B to 4B yields a much larger gain on OCR-centric understanding than on parsing. The parsing teacher scores 96.9745 overall, compared with 96.8277 for the Mix-RL student, a gap of about 0.15 points. On understanding, the teacher scores 88.1 versus the student's 83.2, a gap of about 4.9 points, and leads on all five benchmarks. Within each task, the teacher and student use the same amount of training data, although the teacher specializes in a single domain and the student trains on both.

The transfer reversal in Section~\ref{sec:data_level_analysis} complements this capacity comparison: understanding supervision becomes beneficial as the model develops a stronger parsing foundation.

\begin{takeaway}
{\sffamily\bfseries\textcolor{xiaomiorange}{Takeaway 2.}} Scaling from 0.8B to 4B yields a much larger gain on understanding than on parsing. Together with the emergence of positive transfer, this supports a curriculum that builds parsing before placing greater emphasis on understanding.
\end{takeaway}

\FloatBarrier
\subsection{Training Dynamics: MOPD vs. Mix-RL}
\label{sec:training_dynamics}

\begin{figure}[htbp]
\centering
\includegraphics[width=\textwidth]{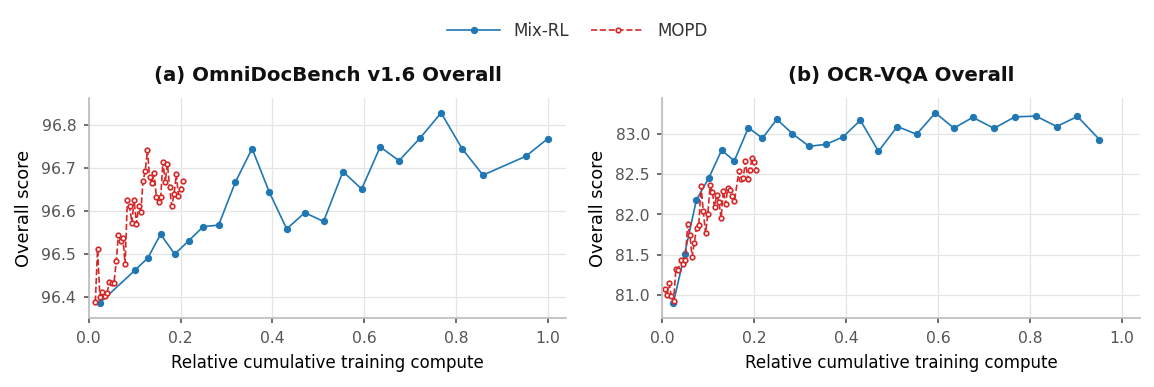}
\caption{Training curves for Mix-RL and MOPD on document parsing and OCR-centric understanding. Both panels report Overall scores; the mean over OCRBench, DocVQA, TextVQA, InfoVQA, and ChartQA measures understanding. Horizontal position indicates cumulative training compute on a shared relative scale.}
\label{fig:rl_vs_mopd}
\end{figure}

Mix-RL scores higher than MOPD on both parsing (96.83 vs.\ 96.74) and understanding (83.2 vs.\ 82.7). Figure~\ref{fig:rl_vs_mopd} compares their training trajectories on a shared relative compute scale. We stopped MOPD training when the monitored scores began to fluctuate; Mix-RL continued longer.

MOPD improves parsing rapidly, reaching 96.7417 at a lower compute budget. The longer Mix-RL run reaches 96.8277. On understanding, the observed peaks are 82.708 for MOPD and 83.264 for Mix-RL. These trajectories show MOPD's rapid early gains in parsing and the higher peaks Mix-RL attains over a longer run.

\begin{takeaway}
{\sffamily\bfseries\textcolor{xiaomiorange}{Takeaway 3.}} During training on OCR-related tasks, MOPD reaches competitive parsing scores at lower compute. At the same time, the longer Mix-RL run attains higher observed peaks on both parsing and understanding.
\end{takeaway}

%% file: tables/analysis_transfer_scale.tex
\begin{table}[htbp]
\centering
\caption{Understanding-to-parsing transfer at three stages of parsing training on OmniDocBench v1.6. Each block shows the successive addition of parsing and OCR-VQA supervision.}
\label{tab:understanding_transfer_scale}
\label{tab:data_order_region}
\label{tab:ocr_20m_vqa}
\label{tab:ocr_64m_vqa}
\footnotesize
\setlength{\tabcolsep}{3pt}
\renewcommand{\arraystretch}{1.15}
\begin{tabular*}{\textwidth}{@{\extracolsep{\fill}}lcccccc@{}}
\toprule
Training data & Overall$\uparrow$ & \makecell{Text\\Edit$\downarrow$} & \makecell{Formula\\CDM$\uparrow$} & \makecell{Table\\TEDS$\uparrow$} & \makecell{Table\\TEDS-S$\uparrow$} & \makecell{Reading Order\\Edit$\downarrow$} \\
\midrule
\multicolumn{7}{c}{\textit{Early stage: 9\% of all parsing data}} \\
Region-level parsing & 93.594 & 0.045 & 96.555 & 88.727 & 92.237 & 0.144 \\
\quad + OCR-VQA & 93.264 & 0.047 & 96.368 & 88.123 & 91.538 & 0.145 \\
\hline
\hline
\addlinespace[4pt]
\multicolumn{7}{c}{\textit{Middle stage: 29\% of all parsing data}} \\
Without page-level data & 94.730 & 0.045 & 96.811 & 91.878 & 94.574 & 0.143 \\
\quad + Page-level parsing & 94.941 & 0.043 & 97.044 & 92.079 & 94.819 & 0.142 \\
\quad + OCR-VQA & 95.580 & 0.039 & 97.536 & 93.103 & 95.512 & 0.140 \\
\addlinespace[4pt]
\hline
\hline
\multicolumn{7}{c}{\textit{Late stage: 50\% of all parsing data}} \\
Region-level parsing & 95.8432 & 0.0345 & 98.3012 & 92.6784 & 95.3894 & 0.1236 \\
\quad + Page-level parsing & 95.8924 & 0.0333 & 98.2048 & 92.8025 & 95.4384 & 0.1232 \\
\quad + OCR-VQA & 95.9673 & 0.0338 & 98.3324 & 92.9495 & 95.4586 & 0.1231 \\
\bottomrule
\end{tabular*}
\end{table}

%% file: tables/analysis_qmask_mixrl.tex
\begin{table}[htbp]
\centering
\caption{Component ablations on OmniDocBench v1.6, with the 0.8B baseline for reference. The Q-Mask and Mix-RL comparisons are evaluated at different training stages.}
\label{tab:stage_wise_omnidocbench}
\footnotesize
\setlength{\tabcolsep}{3pt}
\renewcommand{\arraystretch}{1.15}
\begin{tabular*}{\textwidth}{@{\extracolsep{\fill}}lcccccc@{}}
\toprule
Checkpoint & Overall$\uparrow$ & \makecell{Text\\Edit$\downarrow$} & \makecell{Formula\\CDM$\uparrow$} & \makecell{Table\\TEDS$\uparrow$} & \makecell{Table\\TEDS-S$\uparrow$} & \makecell{Reading Order\\Edit$\downarrow$} \\
\midrule
baseline & 72.21 & 0.186 & 70.728 & 64.491 & 70.010 & 0.179 \\
\midrule
\multicolumn{7}{c}{\textit{Q-Mask ablation}} \\
CPT without Q-Mask & 96.0241 & 0.0378 & 97.5249 & 94.3273 & 96.8657 & 0.1396 \\
\quad + Q-Mask & 96.3198 & 0.0372 & 98.0180 & 94.6613 & 96.9934 & 0.1395 \\
\hline
\hline
\addlinespace[4pt]
\multicolumn{7}{c}{\textit{Mix-RL ablation}} \\
CPT starting checkpoint & 96.4739 & 0.0332 & 98.3444 & 94.3973 & 96.6835 & 0.1221 \\
\quad + Mix-RL & 96.8277 & 0.0313 & 98.5044 & 95.1087 & 97.1929 & 0.1221 \\
\bottomrule
\end{tabular*}
\end{table}

%% file: tables/analysis_rl_omnidocbench.tex
\begin{table}[htbp]
\centering
\caption{Task-specific 4B teachers and 0.8B students on parsing and OCR-centric understanding. The Mix-RL rows report the same model results as the Ours rows in Tables~\ref{tab:omnidocbench_v16_external} and~\ref{tab:ocr_vqa}, respectively. VQA Overall is the arithmetic mean of all five benchmarks on a 0--100 scale. The best value in each column is in bold.}
\label{tab:posttraining_comparison}
\label{tab:rl_omnidocbench}
\label{tab:rl_ocr_vqa}
\footnotesize
\setlength{\tabcolsep}{4pt}
\renewcommand{\arraystretch}{1.15}
\begin{tabular*}{\textwidth}{@{\extracolsep{\fill}}lcccccc@{}}
\toprule
\multicolumn{7}{@{}l}{\textit{(a) Document parsing: OmniDocBench v1.6}} \\
\addlinespace[2pt]
Model & Overall$\uparrow$ & \makecell{Text\\Edit$\downarrow$} & \makecell{Formula\\CDM$\uparrow$} & \makecell{Table\\TEDS$\uparrow$} & \makecell{Table\\TEDS-S$\uparrow$} & \makecell{Reading Order\\Edit$\downarrow$} \\
\midrule
4B parsing teacher & \textbf{96.9745} & \textbf{0.0308} & 98.4102 & \textbf{95.5932} & \textbf{97.5050} & 0.1228 \\
0.8B CPT & 96.4739 & 0.0332 & 98.3444 & 94.3973 & 96.6835 & \textbf{0.1221} \\
0.8B MOPD & 96.7417 & 0.0314 & 98.4677 & 94.8975 & 97.1579 & 0.1223 \\
0.8B Mix-RL & 96.8277 & 0.0313 & \textbf{98.5044} & 95.1087 & 97.1929 & \textbf{0.1221} \\
\bottomrule
\end{tabular*}

\vspace{6pt}
\begin{tabular*}{\textwidth}{@{\extracolsep{\fill}}lcccccc@{}}
\toprule
\multicolumn{7}{@{}l}{\textit{(b) OCR-centric understanding: five OCR-VQA benchmarks}} \\
\addlinespace[2pt]
Model & Overall$\uparrow$ & DocVQA$\uparrow$ & InfoVQA$\uparrow$ & ChartQA$\uparrow$ & OCRBench$\uparrow$ & TextVQA$\uparrow$ \\
\midrule
4B VQA teacher & \textbf{88.1} & \textbf{95.9} & \textbf{84.4} & \textbf{87.1} & \textbf{88.3} & \textbf{84.8} \\
0.8B CPT & 78.1 & 92.2 & 72.5 & 83.4 & 80.6 & 62.0 \\
0.8B MOPD & 82.7 & 93.1 & 74.8 & 84.2 & 83.9 & 77.5 \\
0.8B Mix-RL & 83.2 & 93.1 & 75.1 & 84.6 & 84.6 & 78.6 \\
\bottomrule
\end{tabular*}
\end{table}

%% file: sections/conclusion.tex
\section{Conclusion}
We presented \METHODNAME, a unified 0.8B-parameter vision-language model for document parsing and OCR-centric understanding. An approximately 170M-sample OCR-centric corpus, supported by automated annotation, sample mining, and targeted synthesis, provides broad supervision. Q-Mask-based text anchoring, multi-task continued pretraining, and Mix-RL progressively develop spatial alignment, task coverage, and performance under verifiable task rewards.

The model leads the compared methods on Real5-OmniDocBench with 95.24 and achieves 96.83 on OmniDocBench v1.6 and 87.94 on Wild-OmniDocBench. On OCR-centric understanding, it achieves an average of 83.2 across five OCR-centric benchmarks and 59.09\% accuracy on our in-house KIE benchmark, demonstrating strong performance on both capabilities within a compact model.

Our ablations suggest that positive transfer from understanding to parsing emerges as the parsing foundation develops: the same understanding pool hurts parsing at the early stage but helps at the middle stage, with benefits persisting at the late stage. Using the same task-specific data, the 4B teachers have a larger advantage over the 0.8B Mix-RL model on understanding than on parsing. Together, these findings support a curriculum that establishes parsing before placing greater emphasis on understanding. MOPD converges quickly on parsing, whereas extended Mix-RL reaches higher observed peaks on both capabilities.

Future work will focus on narrowing this understanding gap, improving robustness under physical-world recapture and long-tail script shifts, and extending coverage across document types and languages while preserving a compact inference model.
\newpage

%% file: sections/contributions.tex
\newpage
\section{Contributions and Acknowledgments}
\label{sec:contributions}

All contributors are listed alphabetically by last name.

\noindent
\begin{minipage}[t]{0.45\textwidth}
\textbf{\textsf{\xiaomiblue{Core Contributors}}}
\begin{itemize}[leftmargin=12pt, itemsep=1pt, parsep=0pt, topsep=4pt]
  \item Xin Chen
  \item Anan Du
  \item Feng Feng
  \item Pei Fu\textsuperscript{\dag}
  \item Jian Luan\textsuperscript{\dag}
  \item Longwei Xu
  \item Shaojie Zhang

\end{itemize}
\end{minipage}%
\hfill
\begin{minipage}[t]{0.45\textwidth}
\textbf{\textsf{\xiaomiblue{Contributors}}}
\begin{itemize}[leftmargin=12pt, itemsep=1pt, parsep=0pt, topsep=4pt]
  \item Hang Li
  \item Heng Qu
  \item Cheng Tan
\end{itemize}
\end{minipage}

\let\thefootnote\relax\footnotetext{\textsuperscript{\dag} Corresponding authors: Pei Fu (\texttt{fupei1@xiaomi.com}) and Jian Luan (\texttt{luanjian@xiaomi.com}).}